%% file: root.tex
\documentclass[letter, 10 pt, conference]{ieeeconf}  % Comment this line out if you need a4paper

\IEEEoverridecommandlockouts                              % This command is only needed if 
\usepackage{booktabs}

\usepackage{graphicx} % for pdf, bitmapped graphics files
\usepackage{gensymb}
\usepackage{amsmath} % assumes amsmath package installed
\usepackage{amssymb}  % assumes amsmath package installed
\usepackage{todonotes}

\title{\LARGE \bf
    ML-based tactile sensor calibration: A universal approach
}

\author{
    Maximilian Karl,
    Artur Lohrer,
    Dhananjay Shah, \\
    Frederik Diehl,
    Max Fiedler,
    Saahil Ognawala, \\
    Justin Bayer,
    Patrick van der Smagt
    \thanks{The authors are with the Department of Computer Science, Technische Universit\"at M\"unchen. Patrick van der Smagt is also at fortiss, TUM Associate Institute. Justin Bayer is also affiliated with sensed.io UG (haftungsbeschr\"ankt), M\"unchen, Germany.} %{\tt\small tumman@tacman.de}}%
}

\begin{document}

\maketitle
\thispagestyle{empty}
\pagestyle{empty}

\begin{abstract}
\input{tex/abstract}
\end{abstract}

\input{tex/introduction}
\input{tex/setup}
\input{tex/estimation}
\input{tex/conclusion}
\input{tex/acknowledgements}

\bibliographystyle{plain}
\small
\bibliography{bib/main}{}

\end{document}

%% file: tex/abstract.tex
We study the responses of two tactile sensors, the fingertip sensor from the iCub and the BioTac under different external stimuli.
The question of interest is to which degree both sensors i) allow the estimation of force exerted on the sensor and ii) enable the recognition of differing degrees of curvature.
Making use of a force controlled linear motor affecting the tactile sensors we acquire several high-quality data sets allowing the study of both sensors under exactly the same conditions.
We also examined the structure of the representation of tactile stimuli in the recorded tactile sensor data using t-SNE embeddings.
The experiments show that both the iCub and the BioTac excel in different settings.

%% file: tex/introduction.tex
\section{Introduction}

The use of tactile sensors for robotic hand use has been strongly advocated for in-hand manipulation, haptic exploration, and similar complex tasks.  Indeed, the development of tactile sensors in recent years has been enormous, and some of those sensors are targeted towards robotic hands.

But what must a tactile sensor, used for grasping and manipulation in a robotic setting, be able to do, and how well can it do that?  Since this is, of course task-dependent and then, even when a task is defined, difficult to determine, we suggest to take human manipulation and human tactile sensor use as a touchstone.

Various studies have demonstrated which physical quantities are encoded in the tactile afferents of human fingers for manipulation.  Following the literature on this subject, we distinguish the following three:
\begin{itemize}
  \item \textsl{Slip.}
  Recordings of afferents during actual lifting have revealed that fast adapting (FAI) afferents respond to local slips \cite{Johansson1987:RN269}; this is encoded as vibration.
  
  \item \textsl{Force.}
  When a finger pushes an object, the fingertip is deformed, resulting in different stress profiles.  Mechanoreceptors detect such stress changes; it was correspondingly demonstrated that, amongst others, slow adapting (SAII) nail units provide vectorial information about fingertip forces \cite{Birznieks2009:RN2218}.

   \item \textsl{Curvature.}
Similarly, the response in the majority of tactile afferents to force application are profoundly affected by the curvature of the contact surface \cite{Jenmalm2003:RN1512}. Given that each afferent ``samples'' the local stress this is not unexpected but implies that information about force and curvature is entangled in the afferent population responses.

\end{itemize}

In this paper we decided to quantify two of these three modi in two different tactile sensors. We are using the \emph{BioTac}\cite{fishel2012sensing} sensor having a soft liquid filled surrounding in comparison to the rather hard \emph{iCub} \cite{schmitz2010tactile} finger sensor with a soft but very thin coating. These sensor were chosen from a wide range of available tactile sensors where we refer to good overviews on tactile sensors in the literature \cite{dahiya2010tactile, tegin2005tactile, yousef2011tactile}.
For our analysis we leave out vibration---i.e., slip---detection.  The reasons for this are the following.  First, one of our sensors, the BioTac, is especially designed for detecting vibration for which it contains an additional pressure sensor, delivering data at 2\,kHz.  Not only does that give the BioTac an ``unfair'' advantage over other tactile sensors that do not have this modality built in, but---second---this also makes a fair comparison difficult, as each sensor would require different data processing techniques to indeed quantify vibration.  We would require different data processing techniques for the other sensors, and a subjective comparison would be hard.  It must be remarked, however, that our first analyses have shown that vibration measurement is possible with all sensors that we compared.

In the sequel we will therefore quantitatively compare two different tactile sensors for estimating (a)~force vectors; and (b)~curvature from. 
It will turn out that making use of a linear model suffices for (a), while an ensemble of nonlinear classifiers is necessary for (b).

%% file: tex/setup.tex
\section{Setup}
\label{Setup}
\subsection{Comparison of Tactile Sensors}
\label{Biotacvsicub}
For all our experiments we used one of two tactile sensors, intended for robotic hand use---BioTac and the fingertip sensor for the iCub robot (henceforth misnamed iCub). Even so, there are some key differences between the innards and working of these two sensors and we elaborate on these differences below.

Structurally, the BioTac sensor consists of a rigid core surrounded by an elastic skin which is filled with a conductive fluid. On the other hand, iCub has a solid inner support covered by a flexible PCB \cite{schmitz2010tactile}. The PCB is surrounded by a conductive silicone sheet, which forms the second terminal of the sensor. Due to the elastic nature of the shell, coupled with fluid interface between the skin and the core, BioTac can mimic the softness of the human finger. The small distance between the outer cover and the core in iCub renders it harder to the touch compared to BioTac. Therefore, the touch on the BioTac sensor tends to activate nearby taxels, which is much reduced on the rather hard iCub.

\begin{figure}
	\includegraphics[width=\linewidth,]{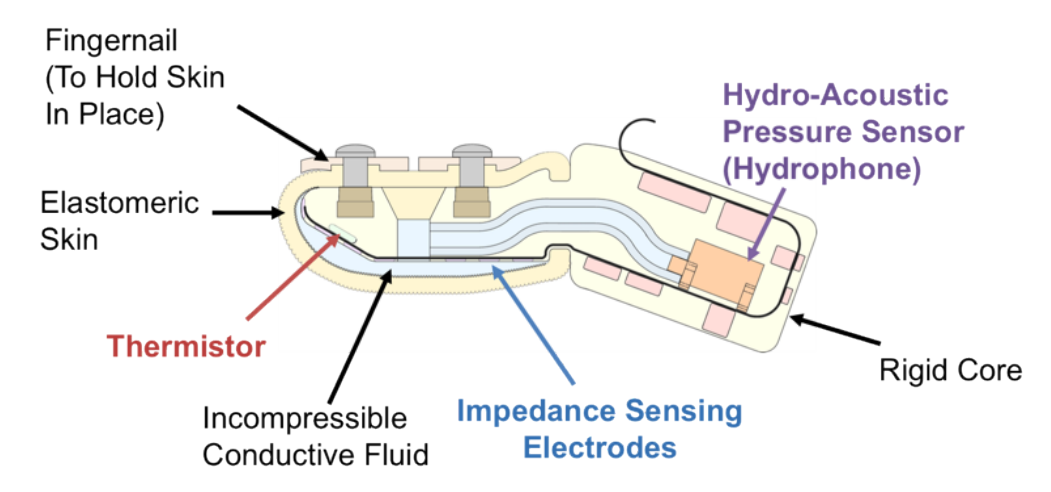}
	\caption{Sideways cross-section of BioTac }
	\label{fig:biotac_side}
\end{figure}

\begin{figure}
	\includegraphics[width=\linewidth,]{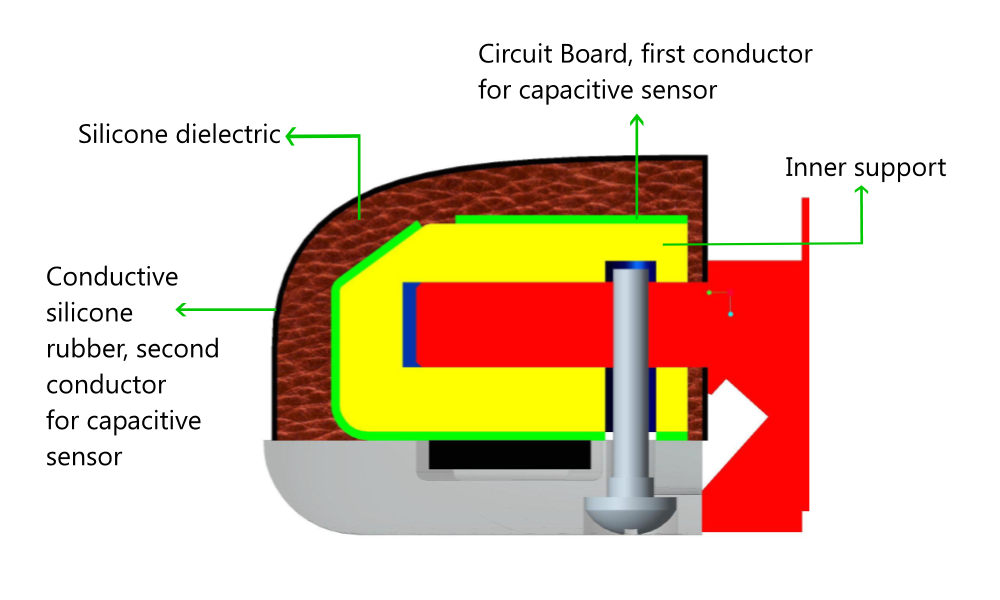}
	\caption{Sideways cross section of the iCub sensor}
	\label{fig:icub_side}
\end{figure}

The tactile sensors in BioTac record impedance values across their electrodes (Fig.~\ref{fig:biotac_side}), while those in iCub record the capacitance values and their change by the way of change in the thickness of the silicone rubber interface (Fig.~\ref{fig:icub_side}). 
As seen from table \ref{tab:sensor_comparison}, while iCub only returns the tactile information from 12 taxels\footnote{\textsl{Taxel} is touch' version of pixel, to be understood as \textsl{tactile element}.}, BioTac returns pressure and temperature, both in DC and instantaneous AC, in addition to 19 taxels. 
\begin{table}[h]
	\caption{Sensor comparison}
	\label{tab:sensor_comparison}
	\centering
	\begin{tabular}{l|c|c}
		& BioTac       & iCub       \\ \midrule
		\multicolumn{1}{l|}{\# taxels}        & 19           & 12         \\ 
		\multicolumn{1}{l|}{measurement principle}           & resistive    & capacitive \\ 
		\multicolumn{1}{l|}{DC Pressure Range}    & 0--100\,kPa    & -          \\ 
		\multicolumn{1}{l|}{DC Temp. Range}       & 0--75$\degree$     & -          \\ 
		\multicolumn{1}{l|}{AC Pressure spectrum} & 10--1040\,Hz   & -          \\ 
		\multicolumn{1}{l|}{AC Temp. spectrum}    & 0.45--22.6\,Hz & -          \\ 
	\end{tabular}
\end{table}

%Hardware
\subsection{Hardware}
\label{Hardware}

\begin{figure}
    \vspace{1em}    % F**KING IEE can't do their templates right
	\includegraphics[width=\linewidth,]{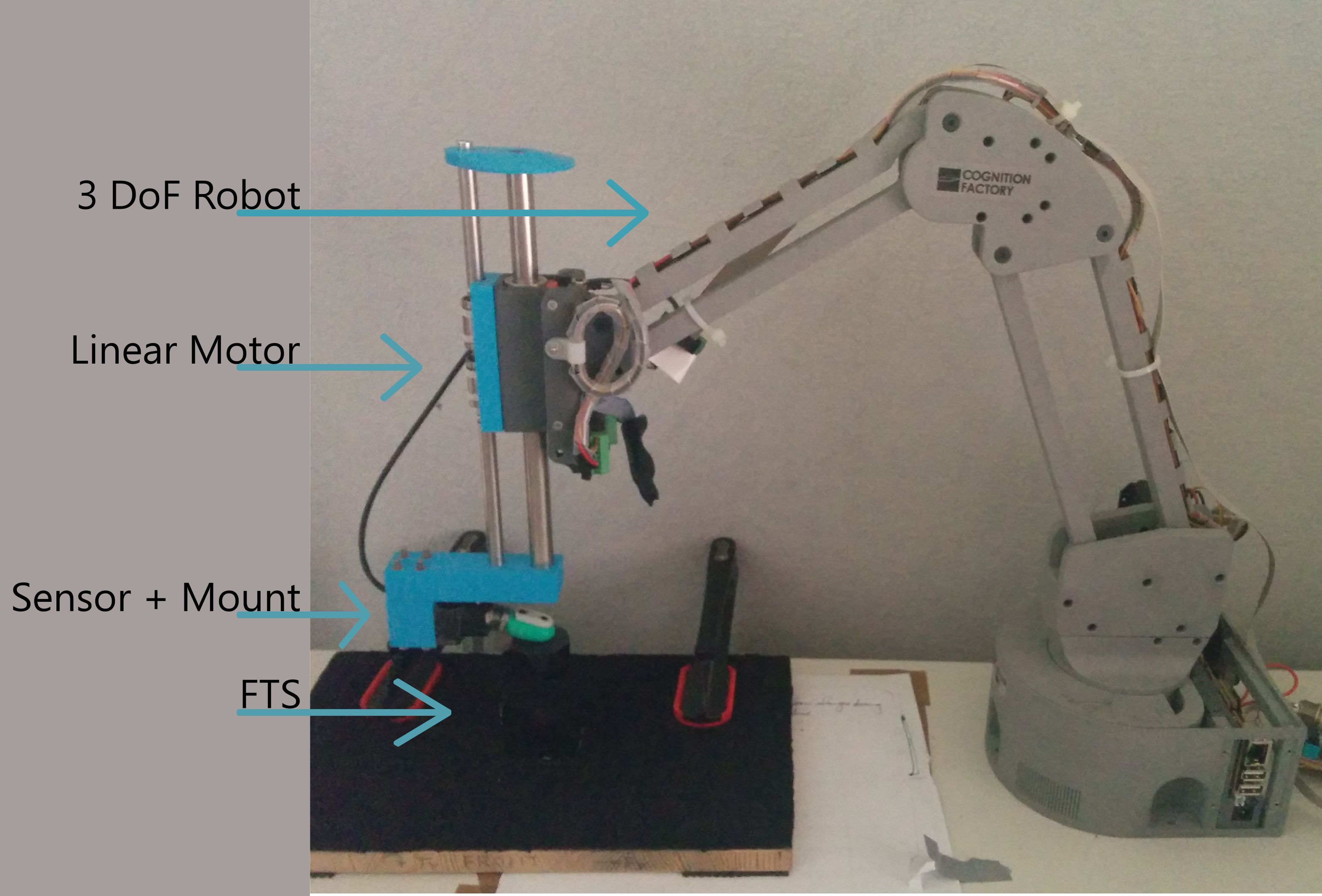}
	\caption{The robot used for performing the experiments in this work.
	}
	\label{fig:setup}
\end{figure}

We performed experiments with a parallel manipulator having three joints and three degrees of freedom (Fig. \ref{fig:setup}). A linear motor acts as an end-effector to the parallel manipulator\footnote{See http://robot-systems.com/.}.
It moves along the $z$-axis which is perpendicular to the common surface of the manipulator and the force-torque sensor.

The force-torque sensor (FTS) used is a Nano-17 manufactured by ATI Industrial Automation. Curvature caps of different radii can be attached to the FTS which is also mounted on the common surface. Mounts were designed to attach the sensors onto the end effector such that the point of contact for both BioTac and iCub remains the same. 

%Electronic Circuitry
\subsection{Electronic Circuitry}
\label{Electronic_circuitry}
The given setup of BioTac, iCub, and FTS are connected to an embedded system using a Serial Peripheral Interface (SPI) bus.
The linear motor is connected to the embedded system using an RS232 serial connection.
The force-torque sensor provides analog data to the Teensy microcontroller where it is then converted to SPI signals using the built-in A/D converter.
BioTac is directly connected to the system using SPI, but the protocol for iCub is converted from Inter-Integrated Circuits (I$^2$C) to SPI.

%Software
\subsection{Software}
\label{Software}
The force exerted by the linear motor on the sensors is controlled using a low pass filtered signal from the force-torque sensor.
The filtering is performed to prevent any oscillations during force control. 
The force recorded for machine learning is the unfiltered force data. 

\subsection{Data Acquisition}
\label{sec:data}

\begin{figure}
    \vspace{.5em}    % F**KING IEE can't do their templates right
	\includegraphics[width=\linewidth,]{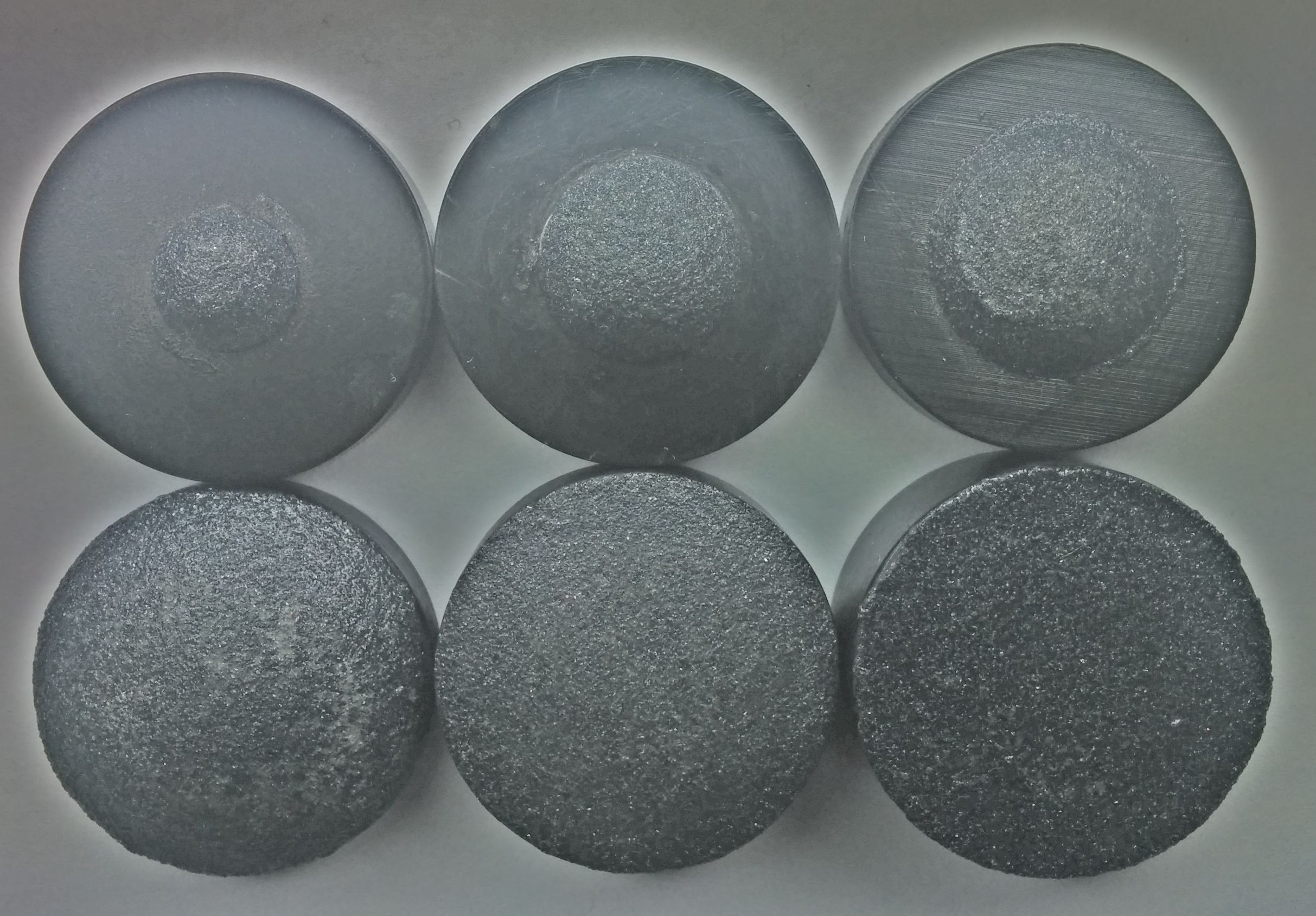}
	\caption{Curved objects touched by the tactile sensors during data acquisition. Radii top row: 5mm, 7.5mm, 10mm. Bottom row: 20mm, 40mm, flat. Photo manipulated for better examination.}
	\label{fig:curvatures}
\end{figure}

We acquired data for two distinct tasks, one regression and one classification task.
In the former, the force in the vertical direction is to be estimated given the raw taxel responses.
The tactile sensor is touching a flat surface in this experiment.
This is different to the classification tasks, where we let the tactile sensors touch one of five curved surfaces (see Fig.~\ref{fig:curvatures}).

During all experiments, the force exerted on the tactile sensors is gradually increased from 0--5\,N until 5000 samples were recorded, resulting in a period of roughly 60 seconds.
Each experiment was performed three times to acquire three distinct data sets for training, validation and testing of our models.

%% file: tex/estimation.tex
\section{Estimation and Recognition}

In all cases for both force estimation and curvature recognitions a minimal but necessary degree of preprocessing was applied.
The first 600 samples were discarded as they featured only the initialization of the setup.
We also performed a simple form of outlier removal: if any of the force or taxel responses violated a ``three sigma rule'', the sample was discarded.
This was done by fitting a Gaussian distribution to a single channel and checking if the difference to the mean exceeded three standard deviations.
Additionally we found it beneficial to apply a ZScore transform individually to each of the sets: each of the training, validation and testing set was divided by its taxel wise standard deviation after subtracting its taxel wise mean.

All experiments were done using the scikit-learn library\cite{scikit-learn}.

\subsection{Force Estimation}
\begin{figure}
    \includegraphics[width=\linewidth,]{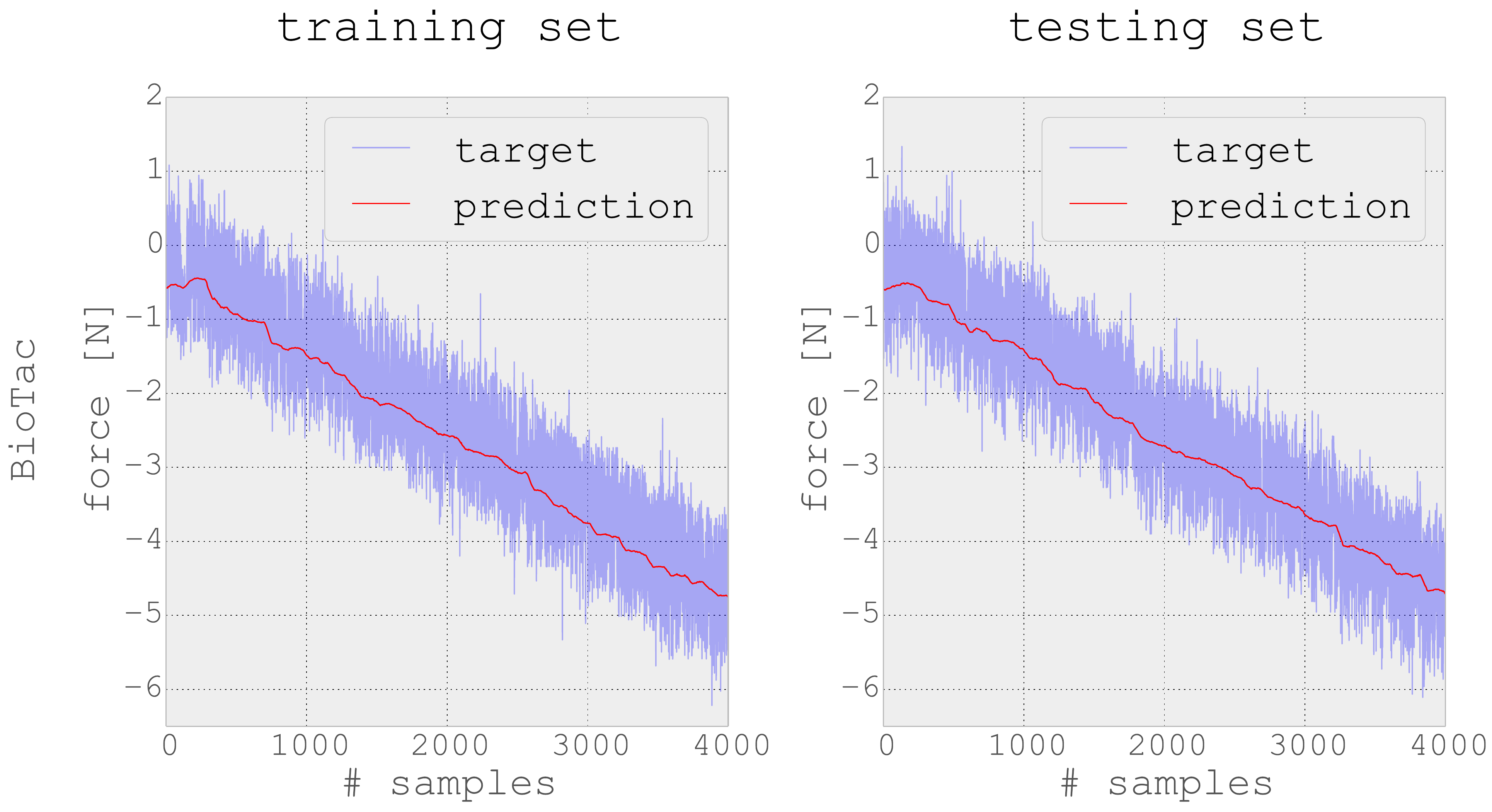}
	\includegraphics[width=\linewidth,]{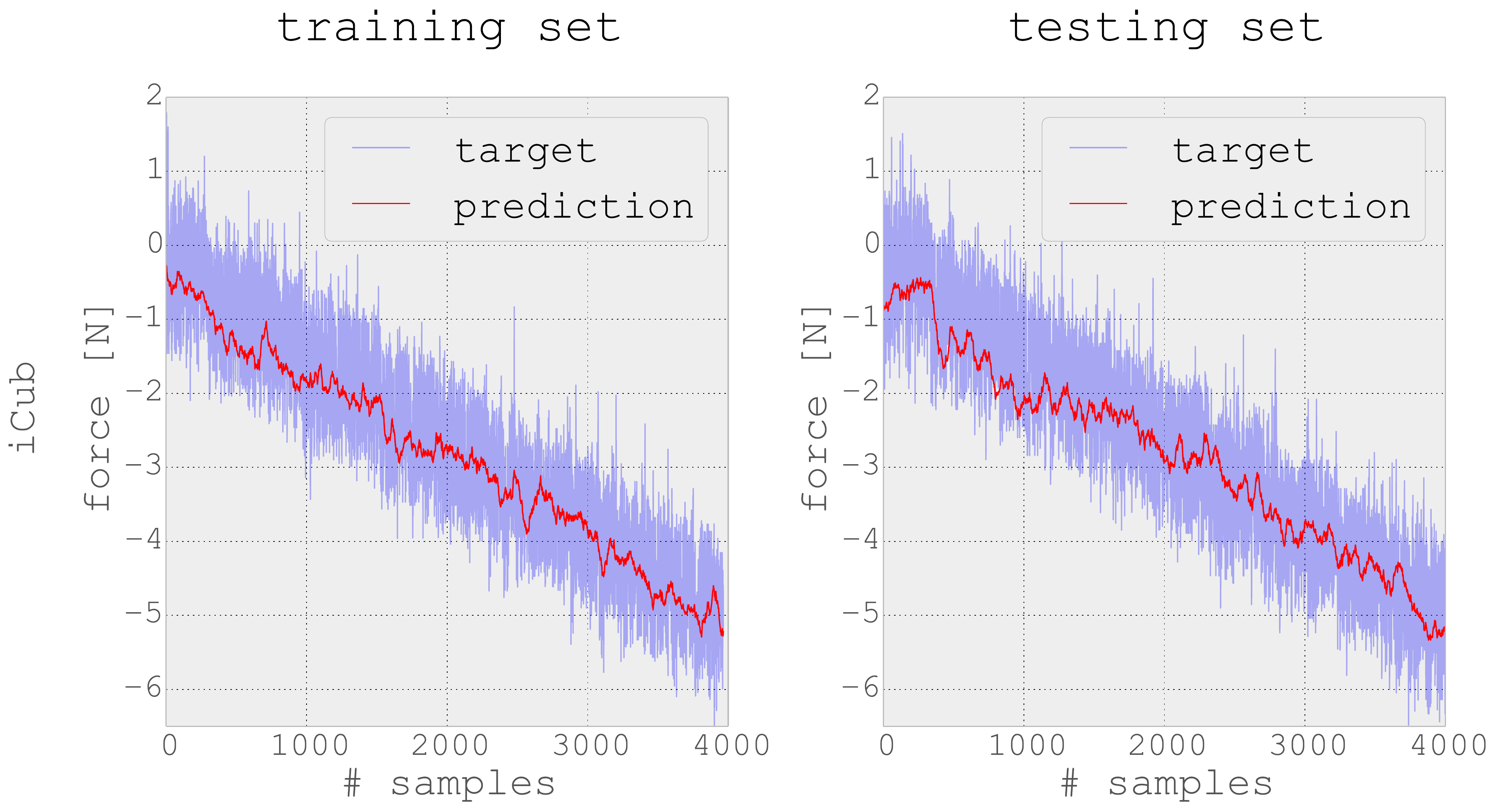}
	\caption{
        We show the results of predicting the force from taxel readings for the iCub (top) and BioTac (bottom).
        The true force readings (blue) and the predictions (red) are shown against time.
        Note that the prediction is is done without giving the regression model input from the force sensor and only tactile data is used.
        The high noise is only coming from a bad signal to noise ration of the force sensor and is not present in the tactile data hence the rather clean prediction.
        In both rows, the left plot corresponds to the training set while the right one corresponds to the test set.
    }
	\label{fig:force-prediction}
\end{figure}

\begin{figure}
    \vspace{1em}    % F**KING IEE can't do their templates right
	\includegraphics[width=\linewidth,]{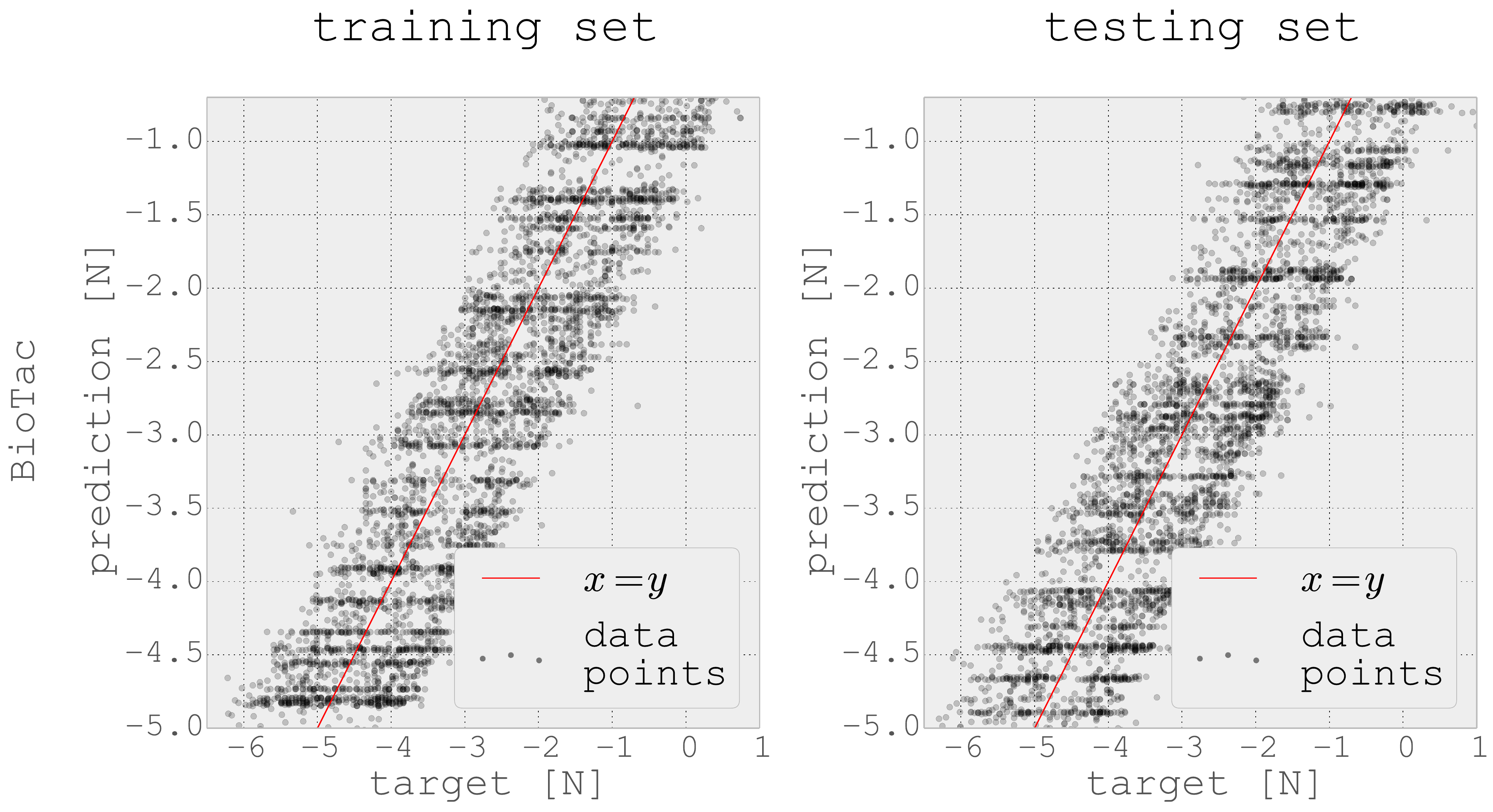}
	\includegraphics[width=\linewidth,]{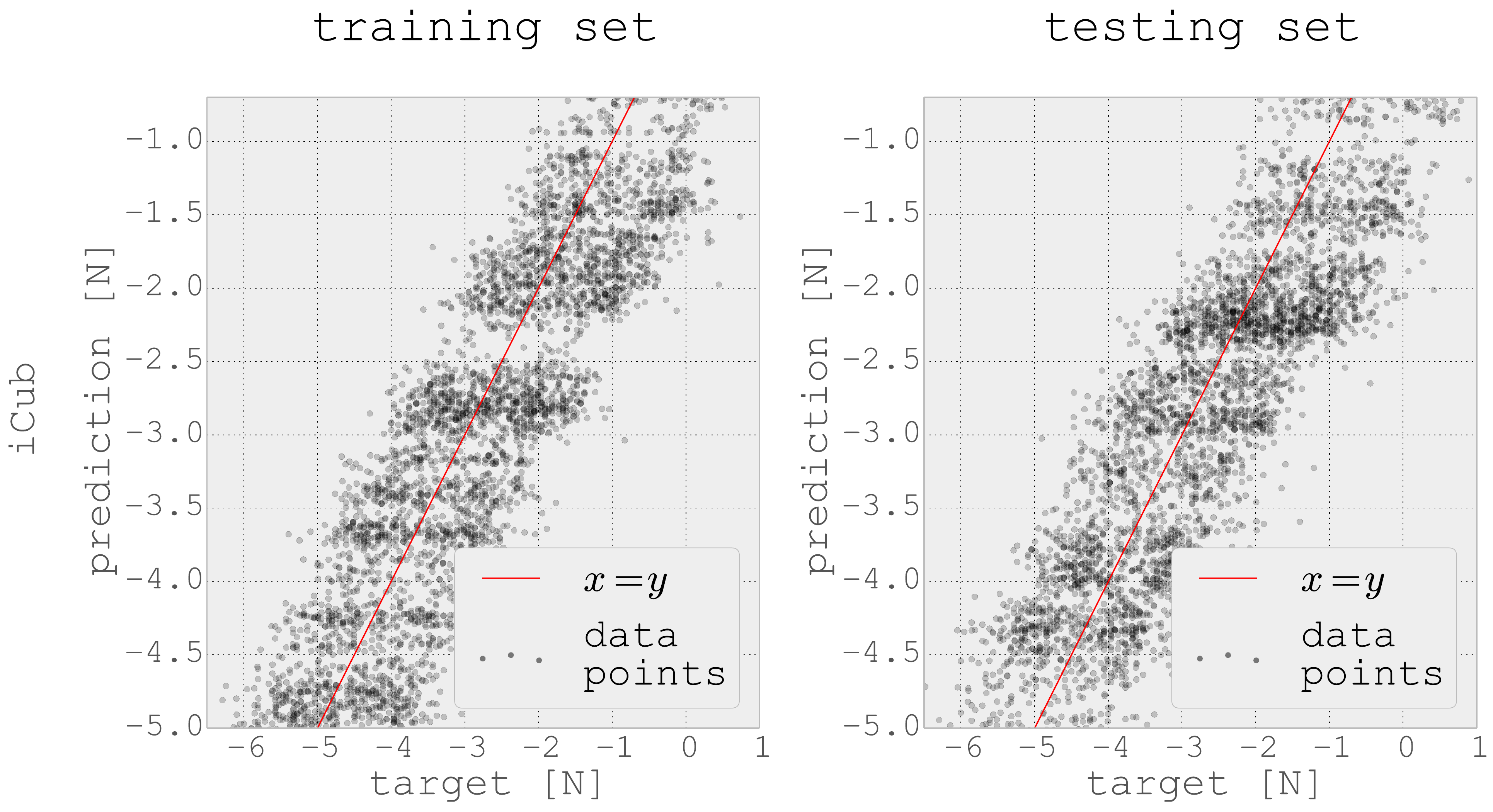}
	\caption{
	    Residual plot for the predictions of the iCub (top) and BioTac (bottom).
        The target force is plotted against the predicted force; perfect predictions would place all predictions on the red line.
    }
	\label{fig:force-residuals}
\end{figure}

Preliminary experiments showed that little was to be gained by employing complex non-linear models, such as random forests or neural networks over a plain linear regression.
We attribute this to the relative simplicity of the task as well due to the little variations in the data.
The only methodological improvement over a simple linear model was to use a sparse linear model, i.e. Lasso (see \cite{murphy2012machine} for a good introductory text).
We attribute this to its smaller sensitivity towards drift, since the drift of the taxels is not incorporated from many but only from few sensors.
To achieve a higher robustness, we performed a window based mean filter, using the sensor readings from time step $t-k$ until $t$ for prediction of the $t$'th force value.
The only hyper parameters $\lambda$ and $k$ were tuned on the validation set and resulted in $\lambda = 0.2$ and $k=20$ for both models.
Parameter $\lambda$ controls the penalisation of the absolute size of coefficients used for regression.
Tuning is done by training the model on the training dataset using different values for these parameters and then testing these models on the validation dataset. The parameter resulting in a model with the best validation error is then selected as the final value.
We achieved a root mean squared error (RMSE) of $0.7307N$ for the BioTac and $0.7659N$ for the iCub.
Note that most of this magnitude is due to the high noise in the (unfiltered) force signal, as can be seen in Fig.~\ref{fig:force-prediction}.
The distribution of the errors is distributed evenly over the whole force range, as can be seen in Fig.~\ref{fig:force-residuals}.
To quantify this, we performed a simple linear regression from the time indices to the force level reflecting the assumption that the force increases linear with time.
These assumptions yielded an RMSE of $0.6875N$, which can be assumed to be a lower bound on the error of a model predicting from taxel responses.

We also note that the predictions of the iCub are much noisier, which is due to the iCub's sensor responses being much noisier as well: the nature of linear models propagates the noise right through to the output.

\subsection{Curvature Recognition}
Preliminary experiments revealed that linear models are not sufficient to separate the data convincingly.
The choice of model was thus mainly driven by its ease of use and reproducibility and ability to model non-linear decision boundaries.
We employed off-the-shelf random forests \cite{breiman2001random} consisting of 500 decision trees each.

\begin{figure}
    \includegraphics[width=\linewidth,]{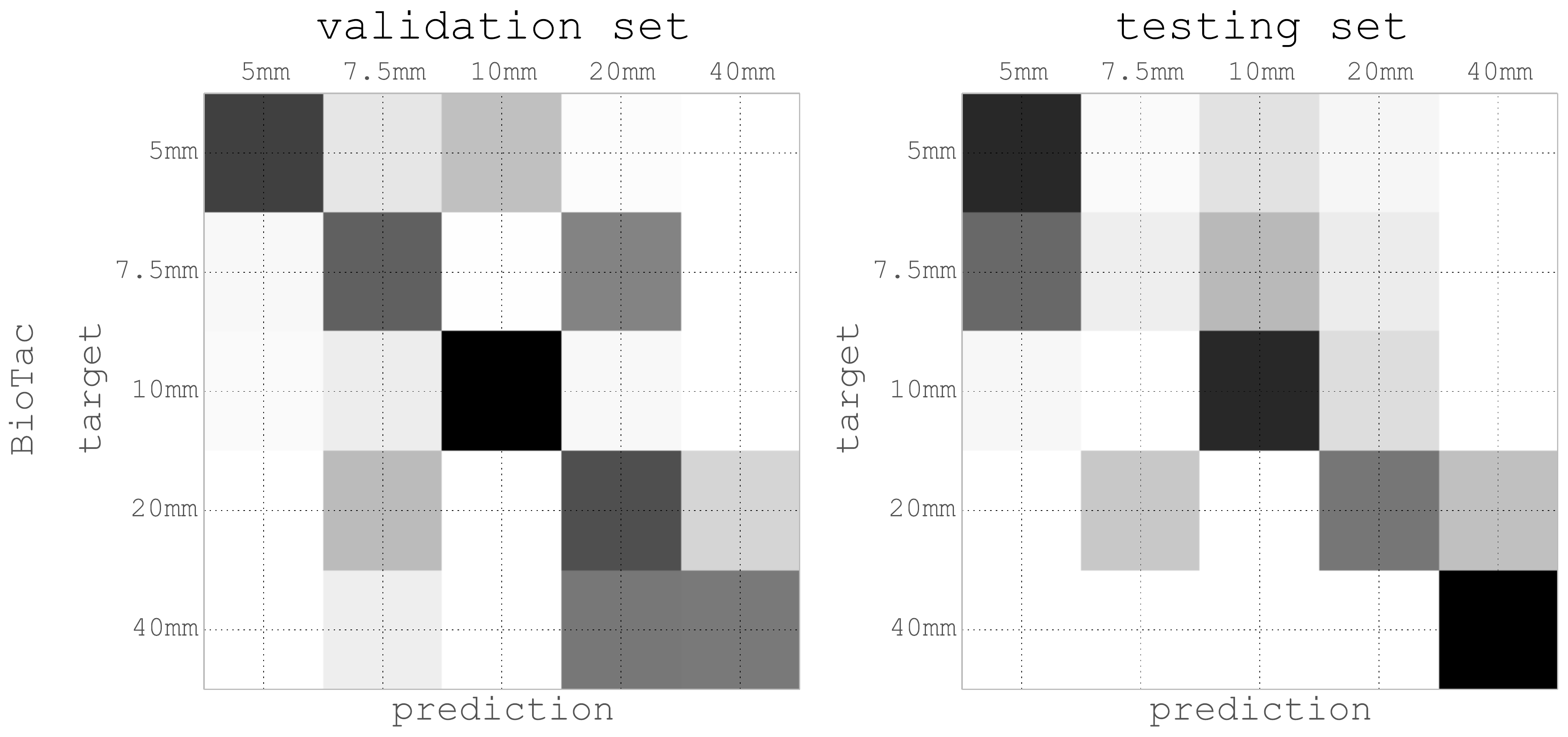}
    \includegraphics[width=\linewidth,]{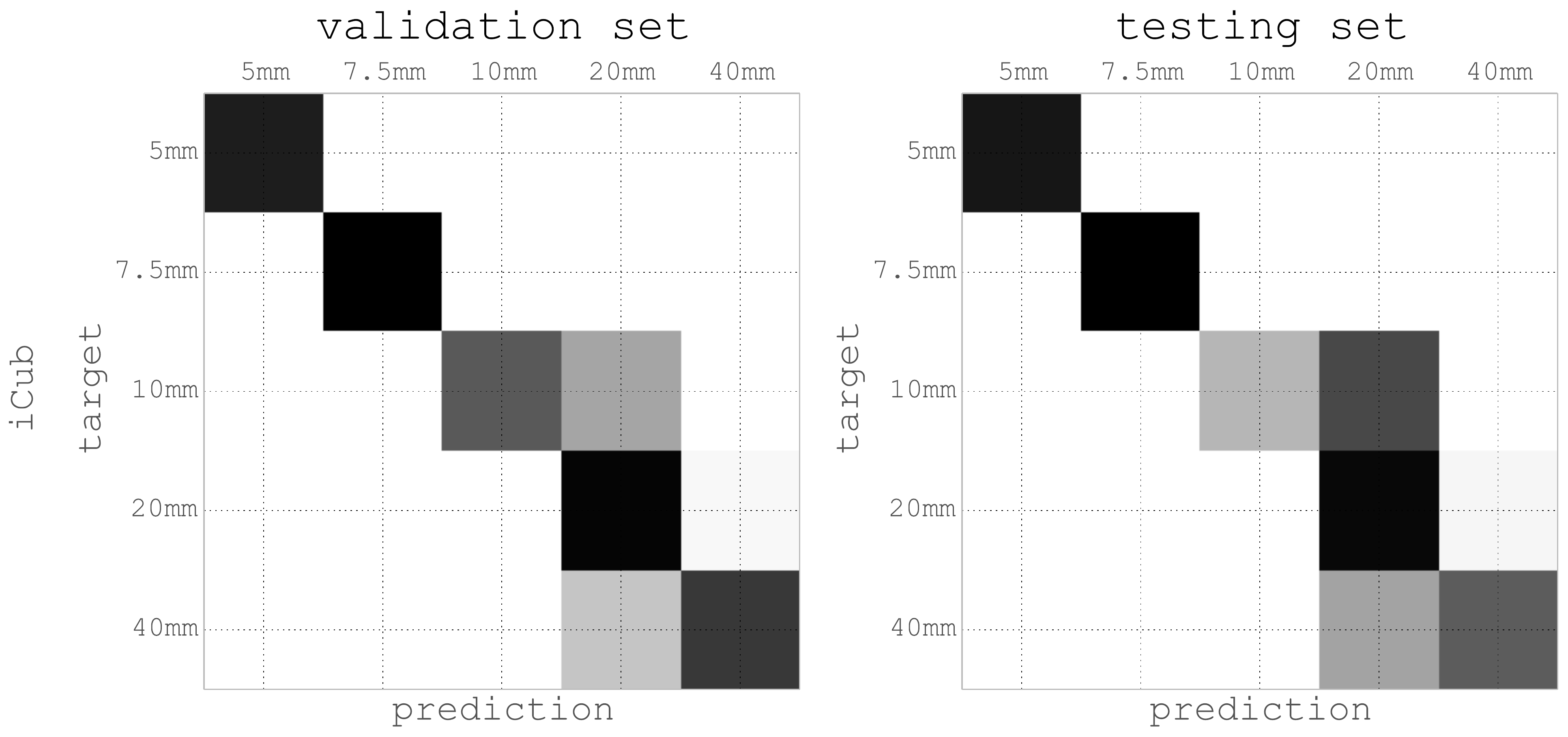}
    \caption{
        Confusion matrices for BioTac (top) and iCub (bottom) on the validation (left) and testing (right) sets.
        Rows represent the target curvatures and columns represent the actual predicted curvatures. For a perfect prediction the matrices would show a diagonal structure. iCub is showing such a diagonal structure indicating better ability to predict curvatures than BioTac
    }
    \label{fig:curvature}
\end{figure}

For evaluation we settled on the F1 score \cite{murphy2012machine} to reflect that we do not have a preference of recall over precision or vice versa.
For the iCub, a F1 score of $0.79512$ was achieved.
The BioTac reached an F1 score of $0.60488$.
Note that random guessing would result in an F1 score of $0.2$.
We show the confusion matrices on the validation and test set in Fig.~\ref{fig:curvature}.
A perfect prediction would show a clear diagonal line in the confusion matrix.
As seen in Fig.~\ref{fig:curvature} iCub is showing such a clear diagonal in contrast to the confusion matrix for BioTac.

\begin{figure}
    \includegraphics[width=\linewidth,]{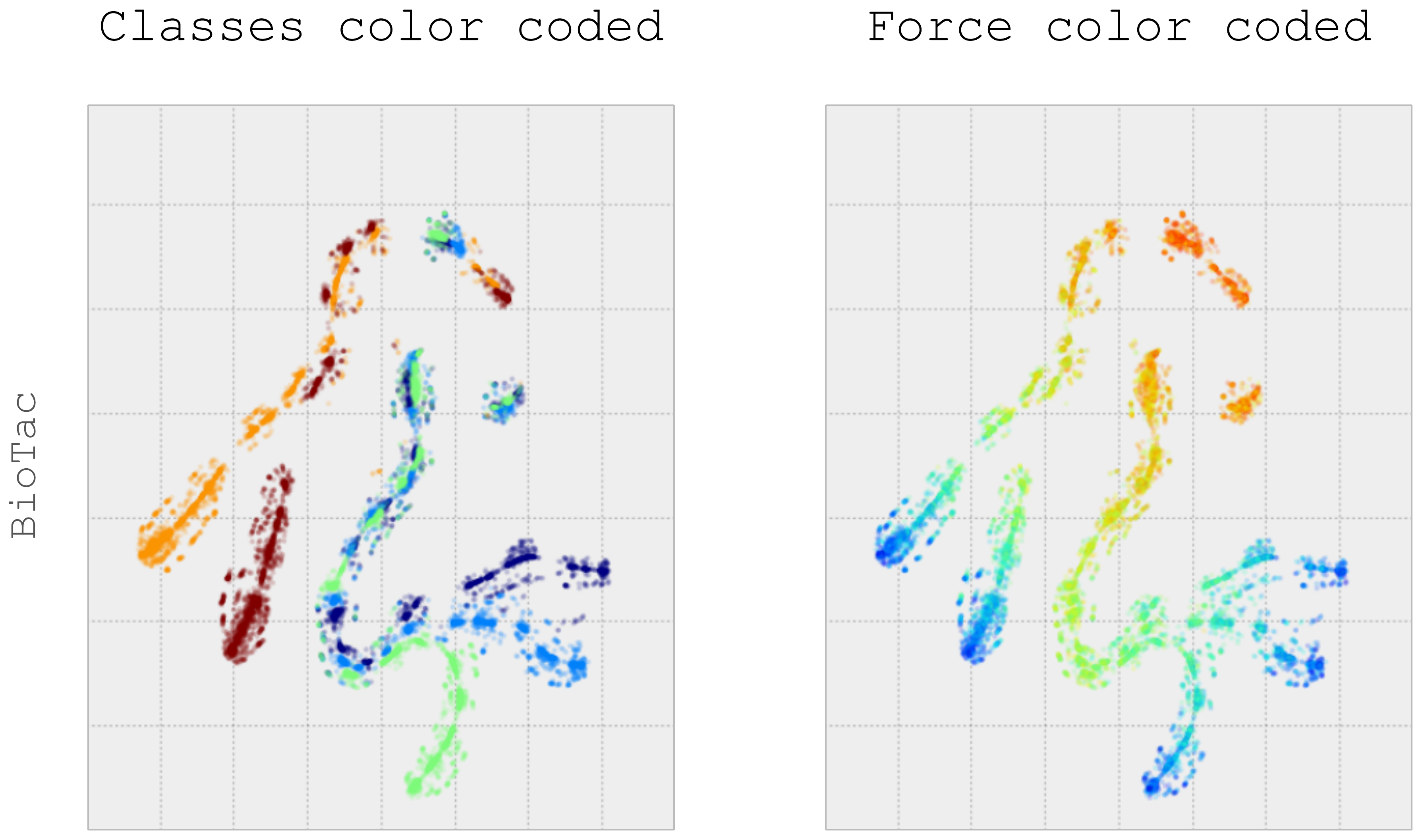}
	\includegraphics[width=\linewidth,]{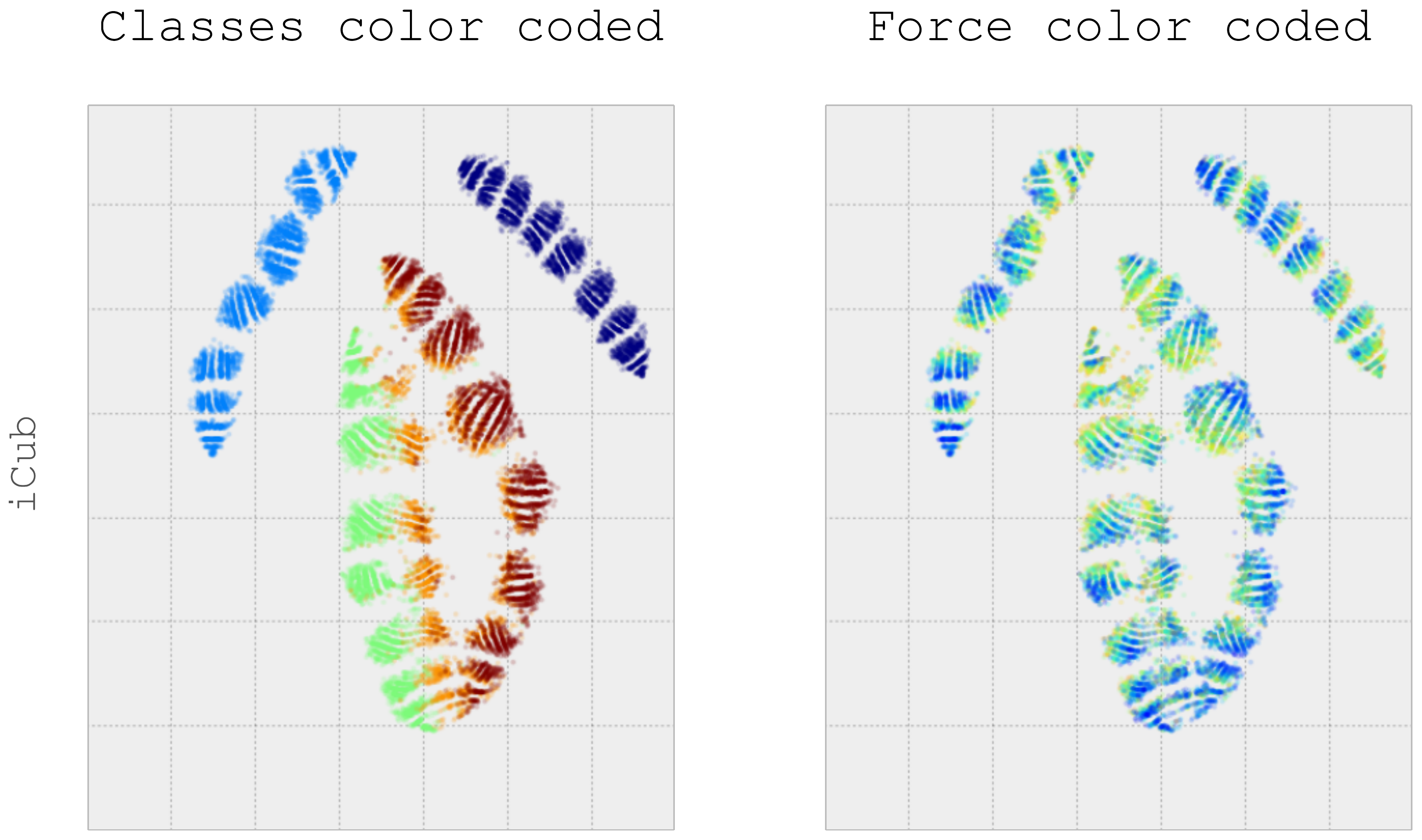}
	\caption{
        t-SNE plots for the iCub (top) and BioTac (bottom), either color coded by class (left) or force level (right).
        Curvature radii are coded as dark blue (5mm), blue (7.5mm), green (10mm), orange (20mm) and red (40mm).
        t-SNE can make hidden structures visible which can otherwise not be visualised due to high dimensionality.
        Red means low force, blue means high force.
        It can be seen immediately that both sensors excel at the opposing task.
        For the iCub, the global structure reflects the differing curvatures while force variations are only consistent locally.
        The BioTac responses are globally consistent with the force levels, but different curvatures can only be separated reliably in the presence of high force.
    }
	\label{fig:tsne}
\end{figure}

We investigated the inferior performance of the BioTac further by making use of a visualisation technique called t-distributed Stochastic Neighbourhood Embedding (t-SNE)\cite{van2008visualizing}.
t-SNE creates 2D-Embeddings out of high-dimensional data in which datapoints are ordered in such a way that close points in the high-dimensional space are also close in the 2D-embedding.
Using these plottings one can gain insight into how the data is structured and which features are separable in tactile space.
We provide evidence that the BioTac's performance increases with force levels, where a better separation is possible.
This can be seen in the t-SNE plots of Fig.~\ref{fig:tsne}. The points in these plots represent the projections of the high dimensional tactile data into a 2D space.
They are colored according to force and curvature values for analysing the different data structure of both sensors.
In the case of the BioTac sensor (top row of Fig.~\ref{fig:tsne}) all curvatures contract to the same tactile stimulation at low forces.
This is not unexpected, since the BioTac's construction distributes low forces over many taxels, making the classification problem harder.
As seen in the bottom row of Fig.~\ref{fig:tsne} iCub is performing better at the opposing task. It encodes curvatures in a global structure while forces are structured locally.
The BioTac sensor data is also showing an order of increasing curvatures from right to left whereas iCub data only represents a local order of curvatures.
Such a clear gradient in the embedding would later help to linearly interpolate unseen curvature radii.

%% file: tex/conclusion.tex
\section{Conclusion and Future Work}
In this work we have studied the responses of different tactile sensors while holding all other variables of the experimental setup constant.
The sensors considered already show substantial difference in their way of representing tactile information.
It is evident that the iCub finger sensor, being relatively stiff and measuring at the surface, encodes curvature globally and force locally.
The BioTac, being soft and measuring ``on the bone'', does quite the contrary by encoding force globally and entangling curvatures in the presence of low force.
While these findings are not unexpected in hindsight, we provided experimental evidence.
Before all, we consider this to be a stepping stone for future evaluations of such hardware: the space of tasks and sensors is yet to be covered.
The tasks considered are remarkably simple and not necessarily representative of those that a tactile sensor empowered robot would face.
E.g., we need to perform similar tests with different force profiles, more diversely formed grounds and more tactile sensors.

%% file: tex/acknowledgements.tex
\section*{Acknowledgements}
Part of this work has been supported in part by the TACMAN project, EC Grant agreement no.\ 610967, within the FP7 framework programme.